\documentclass{article}

\PassOptionsToPackage{numbers, compress}{natbib}

\usepackage[preprint]{neurips_2022}

\usepackage[utf8]{inputenc} 
\usepackage[T1]{fontenc}    
\usepackage{hyperref}       
\usepackage{url}            
\usepackage{booktabs}       
\usepackage{amsfonts}       
\usepackage{nicefrac}       
\usepackage{microtype}      
\usepackage{xcolor}         
\usepackage{graphicx}
\usepackage{amsmath}
\usepackage{xcolor}

\title{Hierarchical Reinforcement Learning for Furniture Layout in Virtual Indoor Scenes}

\author{%
  Xinhan Di\\
  Bloo Company\\
  Shanghai,China\\
  \texttt{xinhan.di@blooxr.com} \\
  \And
  Pengqian Yu \\
  Sea Lab\\
  Singapore\\
  \texttt{yupengqian1989@gmail.com}\\
}

\begin{document}

\maketitle

\begin{abstract}
In real life, the decoration of 3D indoor scenes through designing furniture layout provides a rich experience for people. In this paper, we explore the furniture layout task as a Markov decision process (MDP) in virtual reality, which is solved by hierarchical reinforcement learning (HRL). The goal is to produce a proper two-furniture layout in the virtual reality of the indoor scenes. In particular, we first design a simulation environment and introduce the HRL formulation for a two-furniture layout. We then apply a hierarchical actor-critic algorithm with curriculum learning to solve the MDP. We conduct our experiments on a large-scale real-world interior layout dataset that contains industrial designs from professional designers. Our numerical results demonstrate that the proposed model yields higher-quality layouts as compared with the state-of-art models. 
\end{abstract}

\section{Introduction}
Indoors such as the bedroom, living room, office, and gym are of great importance in people's life. Function, beauty, cost, and comfort are the keys to the redecoration of indoor scenes. Many online virtual interior tools are developed to help people design indoor spaces in graphics simulation. Such virtual simulation provides convenience and a rich experience to customers. Moreover, machine learning researchers make use of the virtual tools to train data-hungry models for the auto layout \cite{Dai_2018_CVPR,Gordon_2018_CVPR}, including a variety of generative models \cite{10.1145/2366145.2366154,10.1145/3303766,Qi_2018_CVPR,10.1145/3197517.3201362}. However, they rely on the external inputs from  interior designers to a large extent and thus are not practical in real-world applications. Therefore, an automatic redecoration method for 3D virtual indoor scenes is in demand.

\begin{figure}
\centering
\includegraphics[width=12cm]{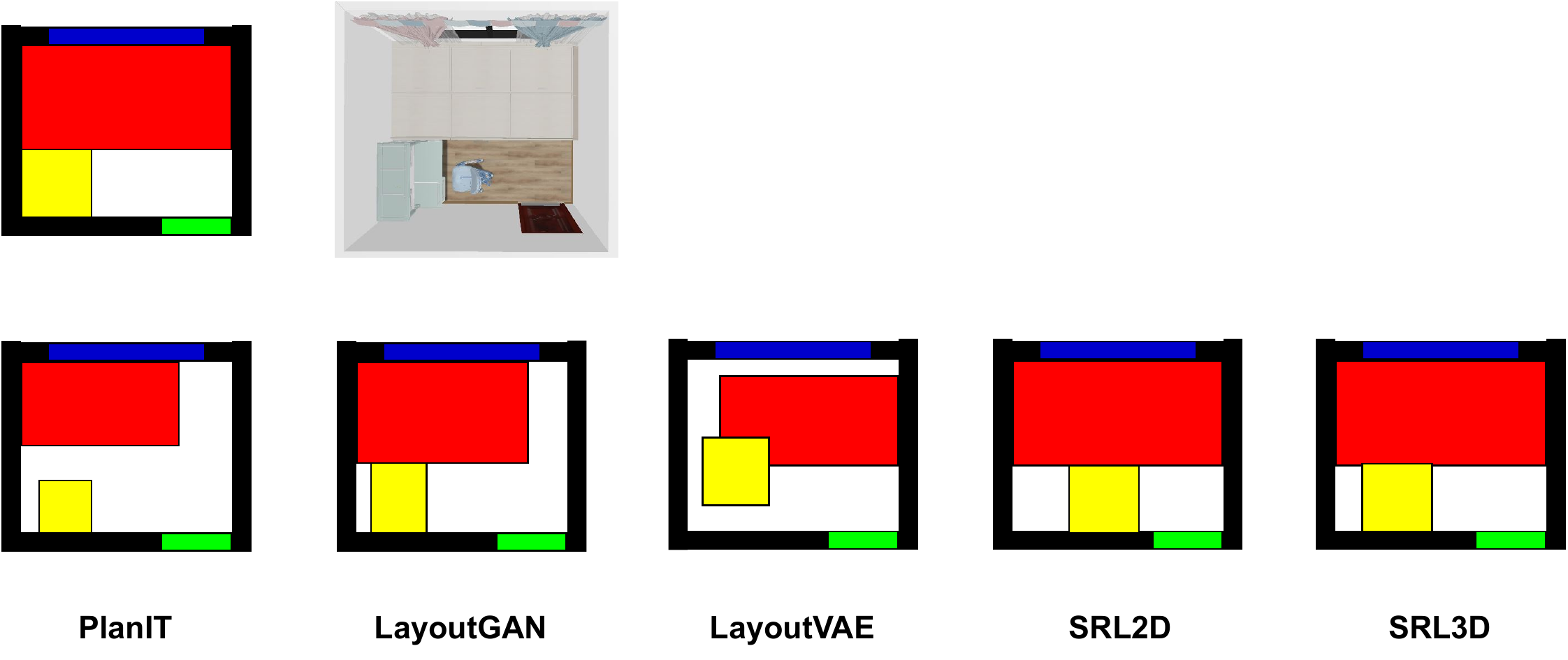}
\caption{An example of two-furniture layout with the state-of-the-art models. The first row represents the ground truth layout in the simulation and 3D render. The second row represents the layouts produced by five state-of-the-art models: PlanIT \cite{10.1145/3306346.3322941}, predicts incorrect size and position of the two furniture. Similarly, LayoutGAN \cite{DBLP:journals/corr/abs-1901-06767} and LayoutVAE \cite{jyothi2019layoutvae} predict incorrect size of position. Besides, both SRL2D \cite{di2021deep} and SRL3D \cite{di2021multi} can only move one furniture.}
\label{fig1}
\end{figure}

Data-hungry methods for synthesizing indoor graphic scenes simulations through the layout of furniture have been well studied. Early work in the scene modeling implement kernels and graph walks to retrieve objects from a database \cite{Choi_2013_CVPR,Dasgupta_2016_CVPR}. The graphical models are employed to model the compatibility between furniture and input sketches of scenes \cite{10.1145/2461912.2461968}. Besides, an image-based CNN network has been proposed to encode top-down views of input scenes \cite{10.1145/3197517.3201362}. A variational auto-encoder is also applied to learn similar simulations \cite{10.1145/3381866,DBLP:journals/corr/abs-1901-06767,Jyothi_2019_ICCV}. Furthermore, the simulation of indoor graphic scenes in the form of tree-structured scene graphs is investigated \cite{10.1145/3303766,10.1145/3197517.3201362,10.1145/3306346.3322941}. {Transformers are applied to play the role of encoder and decoder for the generation of indoor scene \cite{wang2020sceneformer,Paschalidou2021NEURIPS}}. However, as shown in Figure \ref{fig1}. this family of models produces inaccurate size and position for the furniture layout as pointed out in \cite{di2021deep, di2021multi}, 

In \cite{di2021deep, di2021multi}, reinforcement learning (RL; \cite{sutton2018reinforcement, gibney2016google,schrittwieser2020mastering,silver2017mastering})  approaches are proposed to replace the supervised interaction of human designers with the interaction of agents in a designed simulation environment. Specifically,  the industrial interior design process \cite{di2021deep, di2021multi} in the simulated graphic scenes is modeled as a Markov decision process (MDP; \cite{puterman2014markov}), where an agent needs to make multiple decisions for furniture layout in the 3D space by trial and error. To solve the MDP, there are prominent algorithms such as DQN \cite{mnih2013playing} that learns an optimal policy for discrete action space, DDPG \cite{lillicrap2015continuous} and PPO \cite{schulman2017proximal} that train an agent for continuous action space, and A3C \cite{mnih2016asynchronous} designed for a large-scale computer cluster. These algorithms solve stumbling blocks in the application of reinforcement learning in the simulated indoor graphic scenes. 

\begin{figure}
\centering
\includegraphics[width=13cm]{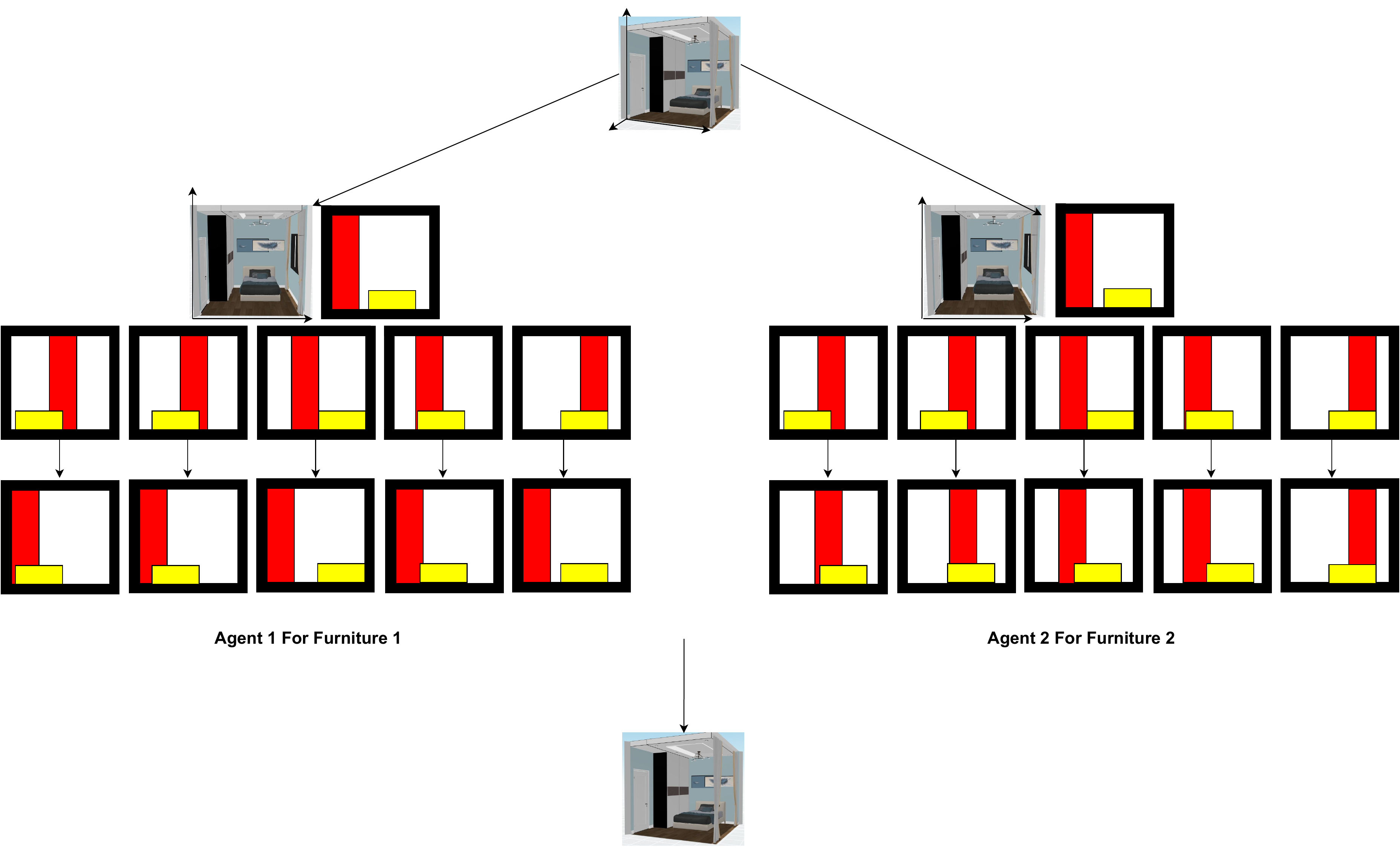}
\caption{An example of two-furniture layout in the virtual indoor scenes. The first row shows the 3D graphic scenes of a room. The second row represents the projected 2D rooms. The third and fourth rows show the movement of two furniture (red and yellow objects). The last row represents the final 3D two-furniture layout in the virtual indoor scenes.}
\label{fig2}
\end{figure}

While the RL approaches proposed in \cite{di2021deep,di2021multi} greatly improve the accuracy of the furniture layout, they are not practical as only one furniture is considered in the planning as shown in Figure \ref{fig1}. Multiple furniture layout remains challenging as one needs to take care of the relative positioning of the furniture. To address this challenge, we formulate the two-furniture layout task as a MDP with hierarchy by defining the hierarchical state, action, and reward function. We then design the simulated environment and deploy hierarchical RL agents to learn the optimal layout for this MDP as illustrated in Figure \ref{fig2}. In particular, we propose to decompose this MDP task through hierarchical training where each furniture layout is considered as a lower-level policy. This hierarchy of nested policies can then be found using the hierarchical actor-critic (HAC) algorithm \cite{DBLP:conf/iclr/LevyKPS19}.

We highlight our two main contributions. First, we formulate the two-furniture interior graphic scenes design task as a Markov decision process problem with hierarchy. Second, we develop an indoor graphic scenes simulator and deploy the hierarchical RL agents which work jointly to learn the two-furniture layout in the simulated graphic scenes. The developed simulator and agents are available at \url{https://github.com/CODE-SUBMIT/simulator3}.

\section{Background and Related Work}

\subsection{Virtual Reality Applications}
There are a variety of recent work on virtual reality applications in the field of sports, business, entertainment, and education. To name a few,
antithetic sampling is developed for Monte Carlo differentiable rendering of games in the entertainment field \cite{zhang2021antithetic}; control strategies are explored for physically simulated characters performing two-player competitive sports \cite{won2021control}; full-body avatars are built through learning-based method for virtual social life \cite{bagautdinov2021driving}; a visible difference predictor is developed for field-of-view video to enlarge the FOV of video in virtual cameras \cite{bagautdinov2021driving}; mixture of volumetric primitives is applied through efficient neural rendering for fast rendering in virtual games \cite{mantiuk2021fovvideovdp}; deep video realistic 3D human character model is developed to display highly realistic shape, motion and dynamic appearance for virtual social life \cite{lombardi2021mixture}; neural network is trained for 3D human motion capturing with physical awareness \cite{habermann2021real}; fast, high-performance, large-scale non-linear least-squares solvers are proposed for efficient calculation in the process of rendering, local mapping \cite{mara2021thallo}, and neural manipulation synthesis with a hand-object spatial representation is developed for convenient interactive method \cite{zhang2021manipnet}. However, to the best of our knowledge, current methods about the application of virtual reality for indoor furniture layout are far from being practical.

\subsection{Indoor Scenes Synthesis}
Our work is related to data-hungry methods for synthesizing indoor graphic scenes simulations through the layout of furniture. Early work in the scene modeling implements kernels to retrieve objects from a database \cite{Choi_2013_CVPR,Dasgupta_2016_CVPR}. The graphical models are employed to model the compatibility between furniture and input sketches of scenes \cite{10.1145/2461912.2461968}. However, these early methods are mostly limited by the size of the scene. It is therefore hard to produce a good quality layout for a large scene size. With the availability of large scene datasets including SUNCG \cite{Song_2017_CVPR}, more sophisticated learning methods are proposed as we review them below.

\subsection{Image CNN Networks}
The image-based CNN network is proposed to encode top-down views of input scenes \cite{10.1145/3197517.3201362}. A variational auto-encoder is applied to learn similar simulations \cite{10.1145/3381866,DBLP:journals/corr/abs-1901-06767,Jyothi_2019_ICCV}, where each column of the 3D arrangement matrix represents an object with location and geometry attributes \cite{10.1145/3381866}. A semantically-enriched image-based representation is learned from the top-down views of the indoor scenes, and convolutional object placement prior is trained \cite{10.1145/3197517.3201362}. {Transformers are applied to build the encoder and  decoder based on attention mechanism \cite{wang2020sceneformer,Paschalidou2021NEURIPS}.} They are used for indoor scene generation with a variety of category of furniture. However, this family of image CNN networks does not apply to the situation where the layout is different with different dimensional sizes for a single type of room.

\subsection{Reinforcement Learning for Furniture Layout}
Authors in \cite{di2021deep} introduce reinforcement learning to model and solve the furniture layout problem. This prior work simplifies the decision-making paradigm in the 3D interior furniture layout task as a MDP problem in a 2D simulation environment. Further work on exploring the 3D layout task is conducted by building two agents for separate two 2D spaces \cite{di2021multi}. However, this simplification is only applicable for a single furniture. In this paper, we extend the problem to a much higher dimensional setting where RL agents need to manage two furniture at the same time. 

\section{Problem Formulation}

\subsection{Markov Decision Process}
Similar to previous work of applying reinforcement learning for the furniture layout \cite{di2021deep, di2021multi}, we are interested in solving a Markov decision process (MDP) augmented with a goal state $G$. A MDP is defined as a tuple $(S,G,A,T,R,\gamma)$, where $S$ is the set of states, $G$ is the goal, $A$ is the set of actions, $T$ is the transition probability function in which $T(s,a,s')$ is the probability of transitioning to state $s' \in S$ when action $a\in A$ is taken in state $s\in S$, $R_t$ is the reward function at time step $t$, and $\gamma\in[0,1)$ is the discount factor. The solution to a MDP is in the form of a control policy $\pi: S,G \rightarrow A$ that maximizes the value function $V_{\pi}(s,g):=\mathbb{E}_{\pi}[\sum_{t=0}^{\infty} \gamma^{t} R_{t}|s_{0}=s,g=G]$ for given initial state $s_0$ and goal $g$.

\subsection{Reinforcement Learning Formulation}

In the two-furniture layout task, the purpose is to efficiently learn a $2$-level hierarchy $\pi$ consisting of $2$ individual policies $\pi_{0}$ and $\pi_{1}$ in parallel. Each $\pi_i$ aims to guide a furniture to its ideal position. This deterministic policy $\pi_i:S_{i},G_{i} \rightarrow A_{i}, i={1,2}$ solves the corresponding MDP $(S_{i}, G_{i}, A_{i}, T_{i}, R_{i}, \gamma_{i})$. Here for each level $i\in\{1,2\}$, the state space is common $S_{i}=S$.

\subsubsection{State}

\begin{figure}
\centering
\includegraphics[width=10.0cm]{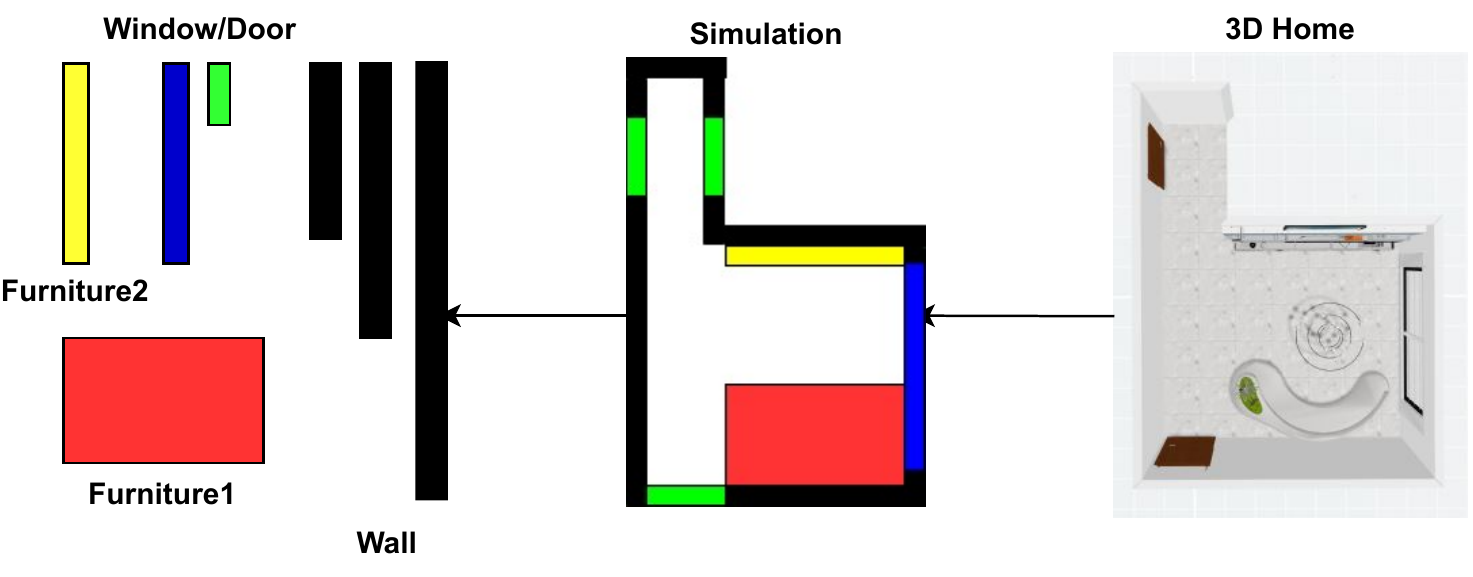}
\caption{Simulation environment for the layout of furniture in the 3D virtual home. The 3D graphic scenes can be decomposed into a 2D simulation scene with components including walls, windows, doors and furniture. For each component, the size and position of each element are represented by its width $w=x_s$, height $h=y_s$, and center $(x,y)=(x_p,y_p)$ of the corresponding box.}
\label{fig3}
\end{figure}

In this paper, the state space $S$ is defined as the geometrical locations of elements such as walls, windows, doors and furniture in the simulation graphic scenes. Each state in the state space consists of the size $s=(x_s,y_s)$ and the position $p=(x_p,y_p)$ of each element as shown in Figure \ref{fig3}.

\subsubsection{Action}
We define the discrete action spaces  $A=(A_{1}, A_2)$ where $A_1$ specifies moving up or down and $A_2$ specifies moving towards left or right, respectively. They jointly describe how the furniture should move to the correct position in each state. As illustrated in Figure \ref{fig2}, agent $1$ works on the first furniture, and agent $2$ works on the second furniture.  Here, moving left denotes the furniture moving towards the $x$-negative direction, while moving right denotes the furniture moving towards the $x$-positive direction. In addition, moving up implies the furniture moving towards the $y$-positive direction, while moving down implies the furniture moving towards the $y$-negative direction.

\subsubsection{Goal and Reward}
The goals $G_{1}$ and $G_{2}$ are defined as the correct positions of furniture in $S_{1}$ and $S_{2}$. The reward function is designed to encourage the furniture to move towards the correct position. For an agent $i$, the reward function has the following form:
\begin{equation*}
    R^i:=\text{IoU}(f^i_{\text{target}},f^i_{\text{state}}),\quad\forall i \in\{1,2\},
\end{equation*}
where $f^i_{\text{target}}$ and $f^i_{\text{state}}$ represent the agent's ground truth and the current layouts for the furniture in the indoor scenes. Here IoU computes the intersection over union between $f^i_{\text{state}}$ and $f^i_{\text{target}}$. 

\section{Hierarchical Actor-critic Framework and Curriculum Learning}

\begin{figure}
\centering
\includegraphics[width=12.0cm]{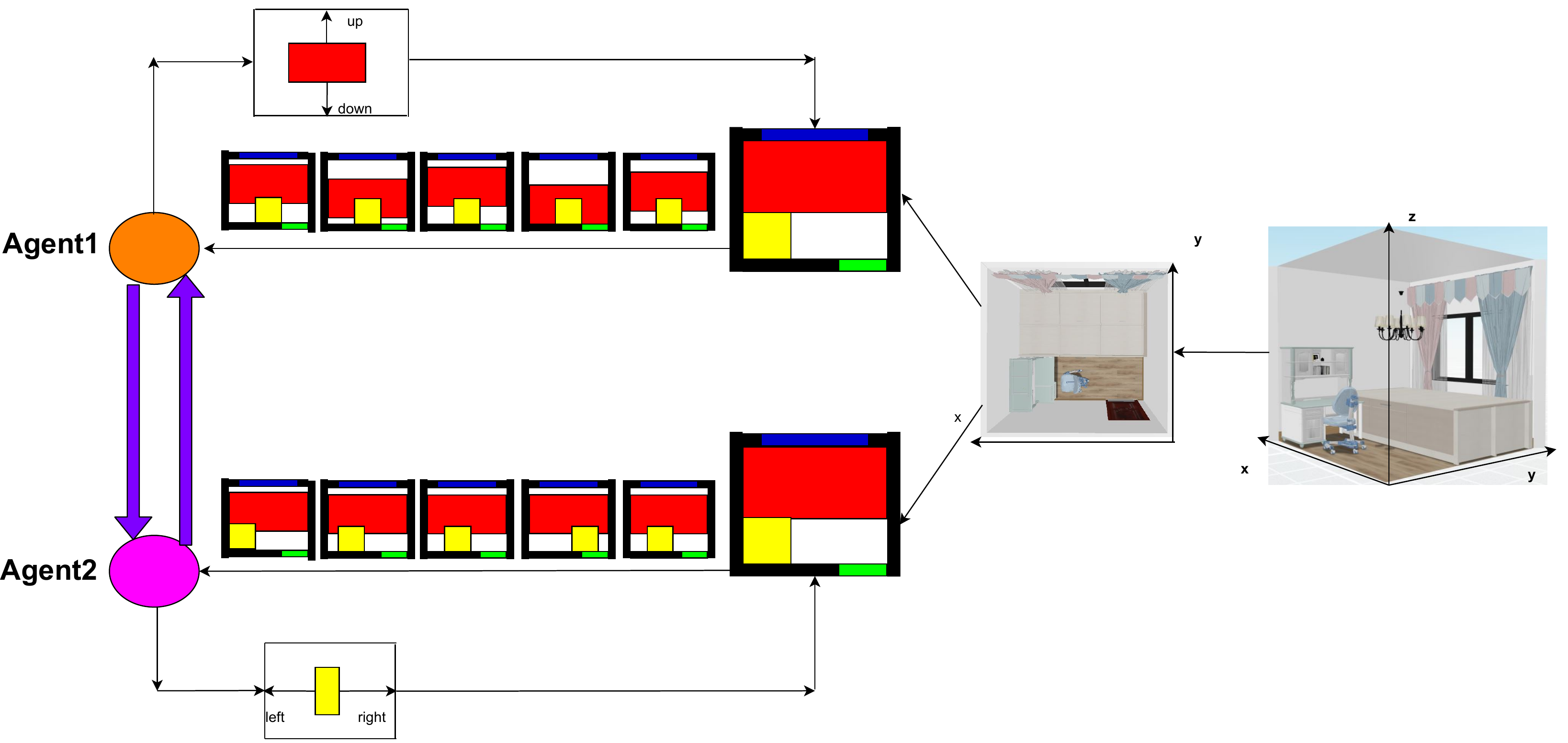}
\caption{The reinforcement learning hierarchy for two-furniture layout in indoor virtual reality. The 3D home is decomposed into 2D scenes with components. Agent $1$ learns a policy of moving furniture $1$ (red object), and agent $2$ learns a policy of moving furniture $2$ (yellow object) simultaneously.}
\label{fig4}
\end{figure}

Hierarchical actor-critic (HAC) module \cite{DBLP:conf/iclr/LevyKPS19}, as shown in Figure \ref{fig4}, is proposed to solve the task of two furniture layout. It overcomes the instability issue that arises when agents try to jointly learn multiple levels of policies in two aspects: $(1)$ a specific hierarchical architecture and $(2)$ a method for learning the different levels of the hierarchy simultaneously and independently. HAC uses nested policies and hindsight transitions, which we discuss in the following.

\subsection{Nested Policies}
HAC module is to learn hierarchies of nested policies. Nesting is critical to decompose problems because it enables agents to learn tasks requiring long sequences of primitive actions. HAC nests policies through embedding the policy at level $i-1$, $\pi_{i-1}$ into the transition function at level $i$ as $T_{i}$. 

The transition function at each sub-goal level, $T_{i}, i>0$ is supposed to work as follows. The sub-goal action $a_{i}$ is assigned to be the goal of level $i-1$, that is $G_{i-1}=a_{i}$, $\pi_{i-1}$ then has at most $H$ attempts to achieve $G_{i-1}$. Here, $H$ is a parameter given by the user. The transition function terminates and the agent's current state is returned when either $\pi_{i-1}$ runs out of $H$ attempts or a goal $G_{n}, n >i-2$ is achieved. Therefore, level $i$'s state transition function $T_{i}$ depends on the full policy hierarchy below level $i$, $\pi_{i-1}$ as from the hierarchy's nested architecture. Similarly, each action from $pi_{i-1}$ depends on $T_{i-1}$, which depends on $\pi_{i-2}$ and so on. Therefore, the notation $T_{i}|\pi_{i-1}$ is used for level $i$'s state transition function for level $i>0$. The notation $T_{0}$ is used to name the base transition function $T_{0}$, and is provided by the task $T_{0}=T$.

\subsection{Hindsight Transitions}
In order to learn multiple policies in parallel, both hindsight action transitions and hindsight goal transitions are applied. The hindsight action and hindsight goal are applied to overcome non-stationary issues which hinder the joint learning of policies. 

Firstly, instead of training each sub-goal policy, we assume a transition function that uses the optimal lower-level policy hierarchy $\pi_{i-1}^{*}$, $T_{i}|\pi_{i-1}^{*}$ is stationary. Such assumption is valid because the function is independent of the changing and exploring lower-level policies, allowing an agent to learn a policy at level $i$ at the same time when the agent learns policies below level $i$.

Secondly, all levels of the hierarchy are supplemented with an additional set of transitions. These are referred as the hindsight goal transitions. This is an extension of the hindsight experience replay \cite{andrychowicz2017hindsight}. The setting is extended as the hierarchical setting, and enables each level to learn more effectively in sparse reward tasks.

Finally, sub-goal testing transitions are implemented for a level to understand whether a sub-goal state can be achieved by the current set of lower level policies. These transitions are implemented as follows. After level $i$ proposes a sub-goal $G_{i}$, the lower level behavior policy hierarchy $\pi_{i-1_b}$ used to achieve sub-goal $a_{i}$ must be the current lower policy hierarchy $\pi_{i-1}$. If the sub-goal $G_{i}$ is not achieved in at most $H$ actions by level $i-1$, level $i$ will be penalized with a low reward. 

\subsection{Training with Curriculum Learning}
Training the proposed HAC module is equipped with curriculum learning in the following two aspects:  $(1)$ we set a curriculum schedule for the goal of the proposed hierarchical actor-critic module; $(2)$ we apply a simple teacher-student strategy in the learning of the curriculum schedule. In particular, we set a curriculum for the IoU goal of furniture $1$ as $\textbf{G}_{1}^{\text{curriculum}}=\{G_{1}^{0},\dots,G_{1}^{c}\}$ where $G_{1}^{c}$ represents the stage $c$ of the curriculum, and the IoU goal for the $c$ stage is $0.45+0.05c$ where $c=0,\dots,10$. $\textbf{G}_{2}^{\text{curriculum}}$ for furniture $2$ is constructed similarly. In addition, we use the weights of networks from $c-1$ stage to be the teacher network to choose suitable initial states for training at the stage $c$. 

\section{Simulation and Training Process}

\subsection{Simulation Environment}
As shown in Figure \ref{fig3} and Figure \ref{fig4}, the indoor simulation environment consists of a simulator $\mathcal{F}$, where $(s^{i}_{\text{next}},R^{i})=\mathcal{F}(s^{i},a^{i})$ and $a^{i}\in A^i$ is the action from the agent $i$ in the state $s^{i}\in S^i$. Here simulator $\mathcal{F}$ receives the action $a^{i}$ and produces the next state $s^{i}_{\text{next}}$ and reward $R^{i}$. In the next state $s^{i}_{\text{next}}$, the geometrical position and size of walls, doors, and windows are not changed, and only the geometrical position of the furniture is updated according to the action. Recall that the action space is discrete, and the center of the furniture can move left, right, up and down at each time step. The simulator $\mathcal{F}$ drops the action if furniture moves out of the room, and the furniture will stay in the same place. 

\subsection{Training Process}
To train the proposed HAC module, a set of configurations is used: Firstly, we use above mentioned curriculum threshold value $0.45+0.05c$ to check whether the sub-goal for level $1$ and level $2$ are achieved at stage $c$. Secondly, two PPO agents \cite{schulman2017proximal} are applied to train the two-level reinforcement learning hierarchy. The discount factor $\gamma$ and the learning rate are  $0.95$ and $0.001$.  If the IoU score for each piece of furniture is over $0.9$, we stop the training of this episode and save the weights of PPO agents. We then start to train the next episode. In total, we train at most $1,000$ episodes for each stage of the curriculum.  

\section{Numerical Experiments}
\begin{figure*}
\centering
\includegraphics[width=11.5cm]{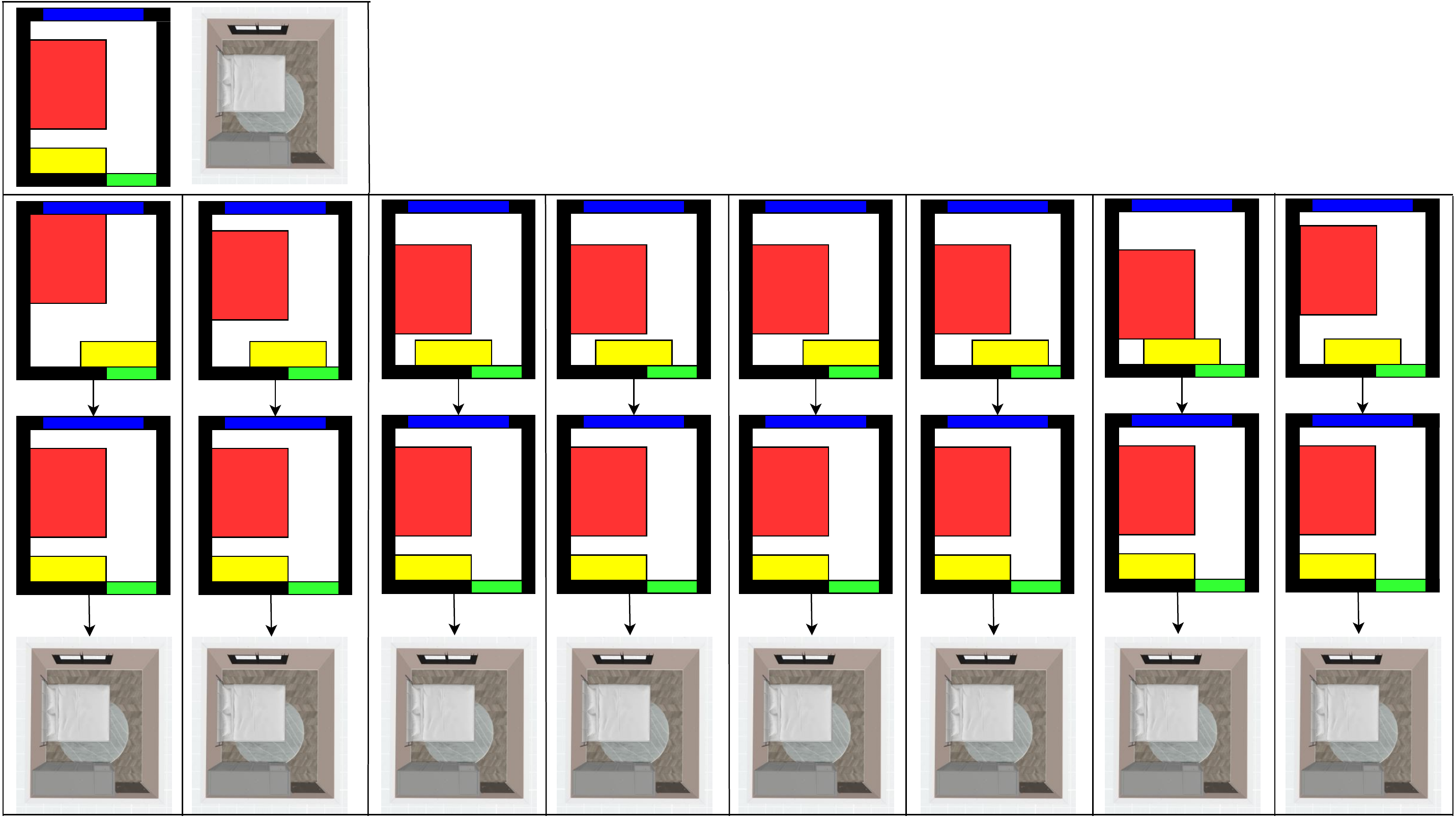}
\caption{Evaluation on the bedroom. The ground truth 3D graphic scene is shown in the first row, and the second row represents the scenes with different initial furniture positions. Here the red objects represent the bed and the yellow objects represent the cabinet. The layouts produced by our method in the simulation are given in the third row. The last row represents the rendering environment in the 3D graphic scenes.}
\label{fig5}
\end{figure*}

\begin{figure*}
\centering
\includegraphics[width=11.5cm]{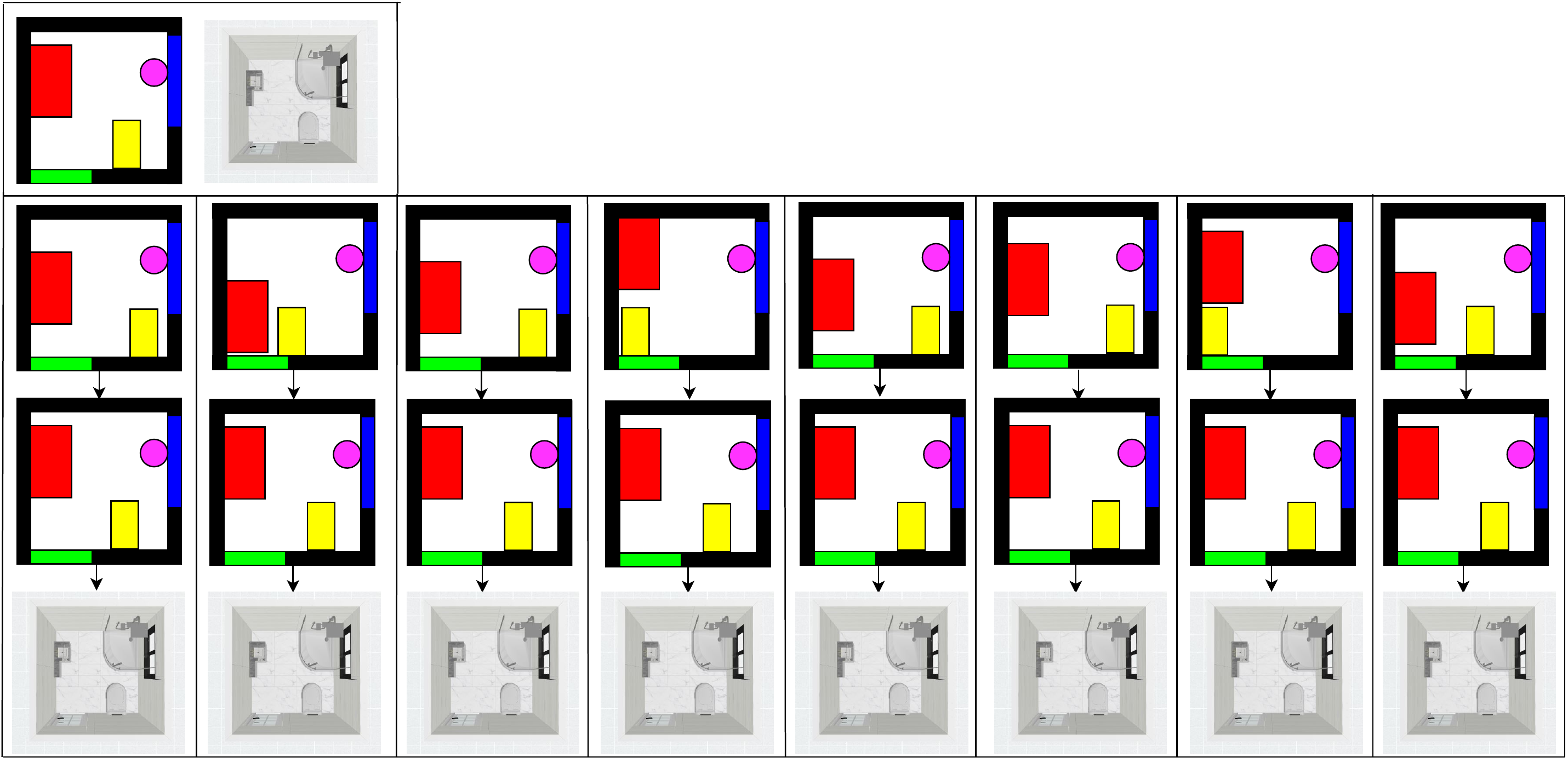}
\caption{Evaluation on the bathroom.  Here the red objects represent the washer and the yellow objects represent the toilet.  }
\label{fig6}
\end{figure*}

\begin{figure*}
\centering
\includegraphics[width=11.5cm]{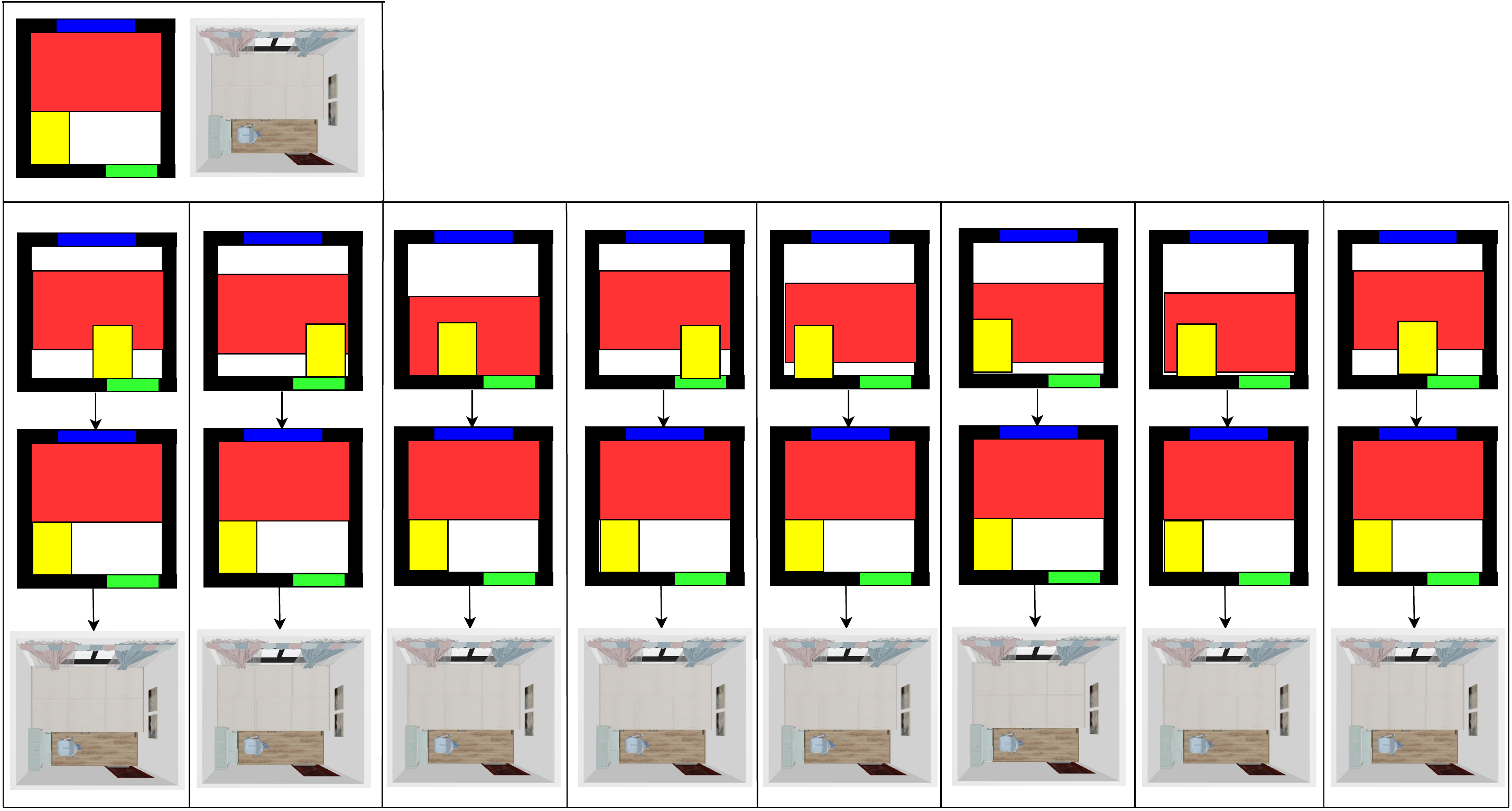}
\caption{Evaluation on the tatami room.  Here the red objects represent the tatami bed and the yellow objects represent the cabinet. }
\label{fig7}
\end{figure*}

\begin{figure*}
\centering
\includegraphics[width=11.5cm]{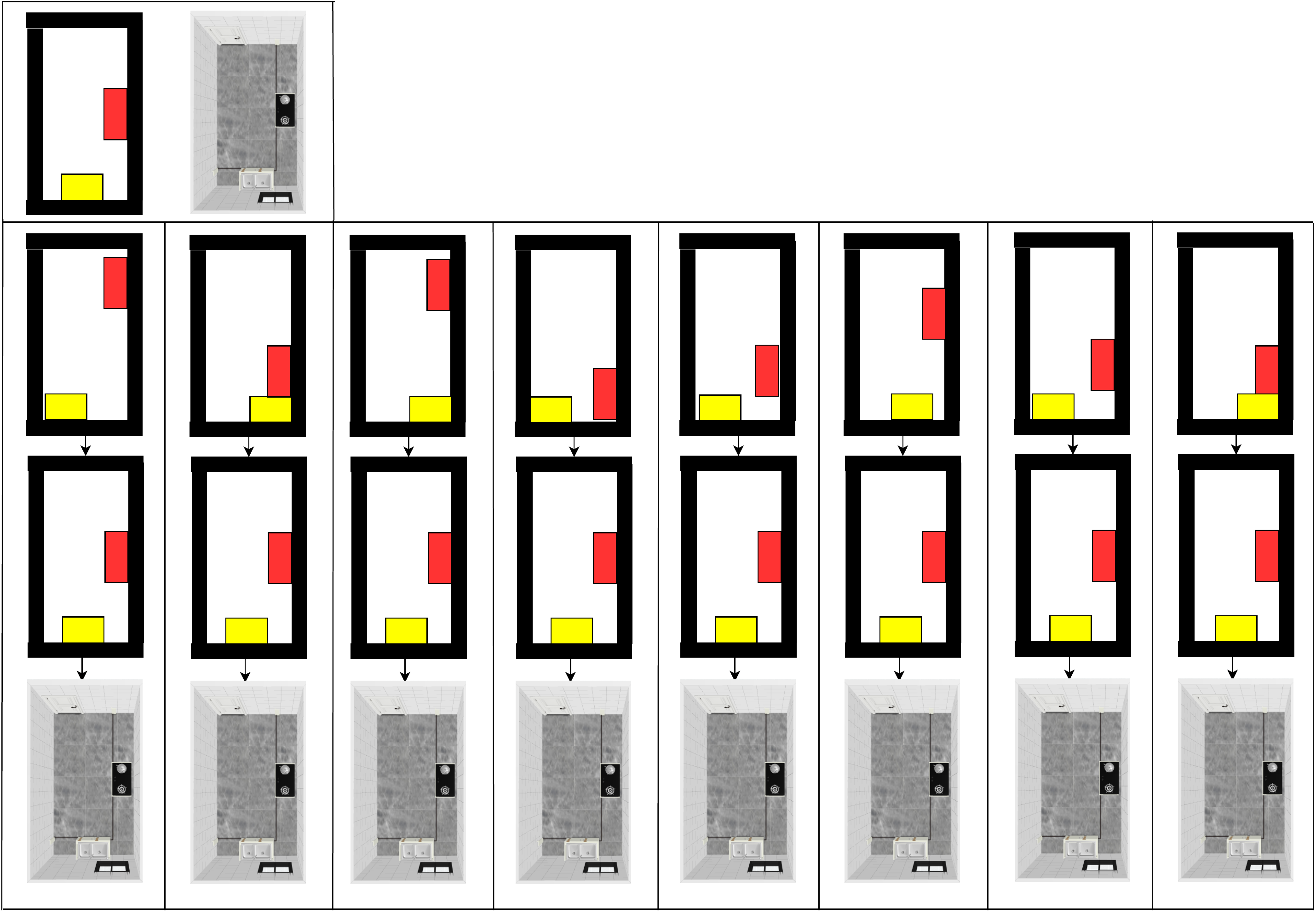}
\caption{Evaluation on the kitchen.  Here the red objects represent the cooker and the yellow objects represent the washer.}
\label{fig8}
\end{figure*}

We discuss the qualitative and quantitative results in this section. In our numerical experiments, four main types of indoor rooms including the tatami, bedroom, bathroom and kitchen are evaluated. In each type of room, there are two types of furniture. For example, there is a tatami bed and a custom cabinet in the tatami room, a toilet and a washer in the bathroom, a normal bed and a custom cabinet in the bedroom, and a cooker and a washer in the kitchen. 

In the simulated environment of each type of room, there are two directions of actions for each type of furniture as shown in Figure \ref{fig4}. In the simulated tatami room environment, actions for the movement of a tatami bed are up and down, and actions for a custom cabinet are left and right. In the simulated kitchen room environment, actions for the movement of a cooker are left and right, and actions for the movement of a washer are up and down. In the simulated bedroom environment, actions for the movement of a bed are up and down, and actions for the movement of a custom cabinet are left and right. In the simulated bathroom environment, actions for the movement of a toilet are up and down, and actions for the movement of a washer are left and down. 

As shown in Figure \ref{fig5}, the agents move the bed (red object), and the custom cabinet (yellow object) from the random starting position. While the bed is moved up or down, the custom cabinet is moved left or right. Eight different initializations are given in the second row. The two RL agents make sequential decisions and move the bed and the customized cabinet hierarchically. The final layouts of the bed and  customized cabinet are shown in the last row. It can be observed that for all eight cases our proposed method results in the expected position. As shown in Figure \ref{fig6} - Figure \ref{fig8}, our proposed method can also produce satisfactory layouts for the tatami room, bathroom, and kitchen. 

In our experiments, we compare our proposed model with the state-of-art baseline methods  include LayoutGAN \cite{10.1145/3306346.3322941}, which uses a novel differentiable wireframe rendering layer that maps the generated layout to a wireframe image, PlanIT \cite{10.1145/3306346.3322941}, which utilizes a graph neural network for the furniture layout and is suitable for 2D simulated layout, and LayoutVAE \cite{jyothi2019layoutvae} which applies a probabilistic and auto-regressive model that generates the scene layout using latent variables in lower dimensions. SceneFormer \cite{wang2020sceneformer} and ATISS \cite{Paschalidou2021NEURIPS} which uses the encoder and decoder based on attention mechanism to generate the indoor scenes. We remark that for a fair comparison with the baseline models, we re-implemented LayoutGAN \cite{10.1145/3306346.3322941}, PlanIT \cite{10.1145/3306346.3322941}, LayoutVAE \cite{jyothi2019layoutvae} SceneFormer \cite{wang2020sceneformer} and ATISS \cite{Paschalidou2021NEURIPS} through applying prior of random position of the furniture. For each room, we test the performance of the proposed model with $2,000$ random starting points. We train on $5,000$ samples for each type of room and test on $1,000$ samples for the corresponding type of room. We use the intersection over union (IoU) to measure the intersection between the predicted layout and the ground truth layout. The comparison for different models is shown in Table \ref{table1} and Table \ref{table2}.

\begin{table*}[h]
\scriptsize
\caption{IoU scores ($\pm$ standard error) for various models (Furniture 1).}
\begin{center}
\begin{tabular}{|c|c|c|c|c|c|c|c|c|}
\hline
\hfil  Room Type    &\hfil ATISS &\hfil SceneFormer &\hfil PlanIT   &\hfil LayoutVAE   & \hfil LayoutGAN  & \hfil Ours \\\hline

\hfil Tatami     &$0.691\pm0.011$ &$0.641\pm0.012$ &$0.651\pm0.009$ &$0.628\pm 0.008$ &$0.645\pm0.013$  &$0.829\pm 0.007$\\\hline

\hfil Bedroom    &$0.672\pm0.015$ &$0.619\pm0.007$ &$0.649\pm0.005$ &$0.632\pm 0.009$ &$0.642\pm0.007$  &$0.819\pm 0.008$\\\hline

\hfil Bathroom   &$0.681\pm0.009$ &$0.627\pm0.011$ &$0.628\pm0.007$ &$0.613\pm 0.008$ &$0.620\pm0.013$  &$0.804\pm 0.009$\\\hline

\hfil Kitchen    &$0.701\pm0.007$ &$0.615\pm0.009$ &$0.634\pm0.012$ &$0.629\pm 0.012$ &$0.631\pm0.013$  &$0.825\pm 0.012$\\\hline

\end{tabular}
\label{table1}
\end{center}
\end{table*}

\begin{table*}[h]
\scriptsize
\caption{IoU scores ($\pm$ standard error) for various models (Furniture 2).}
\begin{center}
\begin{tabular}{|c|c|c|c|c|c|c|c|c|}
\hline

\hfil Room Type    &\hfil ATISS &\hfil SceneFormer &\hfil PlanIT   &\hfil LayoutVAE   & \hfil LayoutGAN  & \hfil Ours \\\hline

\hfil Tatami     &$0.671\pm0.008$&$0.618\pm0.009$&$0.643\pm0.012$ &$0.621\pm 0.009$ &$0.634\pm0.012$  &$0.781\pm 0.006$\\\hline

\hfil Bedroom    &$0.685\pm0.011$&$0.605\pm0.010$&$0.638\pm0.009$ &$0.637\pm 0.012$ &$0.629\pm0.013$  &$0.790\pm 0.009$\\\hline

\hfil Bathroom   &$0.691\pm0.009$&$0.623\pm0.008$&$0.615\pm0.008$ &$0.623\pm 0.011$  &$0.619\pm0.009$ &$0.786\pm 0.011$\\\hline

\hfil Kitchen    &$0.687\pm0.015$&$0.619\pm0.013$&$0.627\pm0.011$ &$0.625\pm 0.009$ &$0.616\pm0.015$  &$0.793\pm 0.008$\\\hline

\end{tabular}
\label{table2}
\end{center}
\end{table*}

\begin{table*}[h]
\scriptsize
		\caption{Percentage ($\pm$ standard error) of 2AFC perceptual study for various models where the real sold solutions are judged more plausible than the generated scenes.}
		\begin{center}
			\begin{tabular}{|c|c|c|c|c|c|c|}
				\hline
				Room Type &ATISS &SceneFormer &PlanIT &LayoutVAE &LayoutGAN & Ours \\\hline
                Tatami&$73.18\pm4.19$&$83.51\pm3.17$&$87.29\pm5.26$&$86.31\pm5.31$&$87.25\pm6.47$&$62.58\pm5.69$\\\hline
                Bedroom&$71.34\pm3.26$&$81.36\pm3.69$&$84.31\pm6.25$&$83.69\pm5.94$&$84.91\pm5.82$&$63.27\pm5.72$\\\hline
                Bathroom&$76.29\pm2.57$&$82.47\pm3.91$&$87.12\pm3.62$&$82.69\pm5.82$&$85.82\pm5.61$&$62.18\pm4.48$\\\hline
                Kitchen&$74.14\pm3.41$&$83.71\pm3.42$&$86.57\pm5.19$&$82.74\pm5.91$&$84.91\pm5.72$&$63.85\pm4.91$\\\hline
			\end{tabular}
		\label{table3}
		\end{center}
\end{table*}

Similar to PlanIT \cite{10.1145/3306346.3322941}, we also conduct a two-alternative forced-choice (2AFC) perceptual study to compare the images from generated scenes with the corresponding scenes from the sold industrial solutions. The generated scenes are produced by our models, PlanIT \cite{10.1145/3306346.3322941}, LayoutVAE \cite{jyothi2019layoutvae} and LayoutGAN \cite{DBLP:journals/corr/abs-1901-06767}, SceneFormer \cite{wang2020sceneformer} and ATISS \cite{Paschalidou2021NEURIPS}  respectively. Ten professional interior designers were recruited as the participants. The scores are shown in Table \ref{table3}. It can be concluded that our proposed approach outperforms the state-of-art models.  

\section{Conclusion}

In this paper, we tackle the interior decoration problem with furniture layout in virtual reality. We formulate the problem as a hierarchical reinforcement learning task. We further solve the problem of two-furniture design and produce a high-quality layout. It is worthwhile to extend our work and consider planning for different shapes of rooms. The auto layout of furniture for multiple rooms or the entire house is also a promising direction to be explored in virtual reality.

{
\bibliographystyle{ieee_fullname}
\bibliography{egbib}
}

\end{document}